\documentclass[11pt]{article}
\usepackage{acl}
\usepackage{times}
\usepackage{latexsym}
\usepackage{amsmath}
\usepackage{amssymb}
\usepackage{graphicx}
\usepackage{url}
\usepackage{booktabs}
\usepackage{multirow}
\usepackage{enumitem}
\usepackage{float}

\title{FOLBench: A Logic-Grounded Evaluation Suite for Evaluating Open-Ended Responses using Graph-Based Semantic Metrics on First-Order Logic form of Natural Language}

\author{
  Gosuddin Kamaruddin Siddiqi\\
  \texttt{Univeristy of Washington, Seattle}\\
  \texttt{gosuddin@uw.edu}
}

\begin{document}
\maketitle
\begin{abstract}
Evaluating open-ended responses from humans or large language models (LLMs) remains a fundamental challenge. Existing metrics—ranging from BLEU \cite{papineni2002bleu} and ROUGE \cite{lin2004rouge} to embedding-based similarity \cite{zhang2020bertscore,zhao2019moverscore,sellam2020bleurt} and LLM-as-a-judge \cite{zheng2023judgelm,chiang2023canllmsjudge}—struggle to capture logical fidelity, structural consistency, and semantic entailment. The outcome of this paper entails \textbf{FOLBench}, a logic-grounded evaluation suite that consists of graph based metrics built from parsing the First-Order Logic (FOL) translated from Natural Language (NL). The suite of symmetric metrics and its composition compares the First-Order Logic forms of candidate Natural Language and gold references. FOLBench provides interpretable, deterministic, verifiable, and compositional evaluation signals. A parameter efficient fine-tuned small language model is also shared to facilitate translations from Natural Language (NL) to First-Order Logic (FOL). In addition, this fine-tuned language model is quantized using AWQ quantization method, thereby minimizes the high VRAM requirement to generate First-Order Logic. Experiments show that FOLBench exposes reasoning failures invisible to existing metrics and correlates strongly with expert human judgments. The evaluation suite is available as pip package\footnote{PyPI: https://pypi.org/project/folbench/, alternatively can work with `pip install folbench`}, and the source code is open to further contribution and reproducibility at GitHub \cite{siddiqi2026folbench}. The fine-tuned models are hosted at HuggingFace space \cite{gosuddin_kamaruddin_siddiqi_2026,gosuddin_kamaruddin_siddiqi_2026_merged}.
\end{abstract}

\section{Introduction}

The pace at which content is created or generated by LLMs has significantly obscured the distinctions against Golden Standards, rubrics, policies or guidelines. The content may exhibit surface-level fluency that is likely to introduce a bias in human judges to trust grammatical perfection and professional tone for true factual accuracy. \cite{akpinar2026llm} Since LLMs mirror human sentence patterns perfectly to generate expected answers thereby hiding a lack of actual conceptual logic. \cite{liu2025applied}  As document length increases, human fatigue compounds exponentially, causing reviewers to miss fine-grained errors.\cite{brunye2026human,gu2026survey}

Large language models increasingly operate in settings where correctness depends not only on linguistic fluency but also on logical coherence and structural consistency. However, the evaluation methods at scale remain dominated by surface-form metrics such as BLEU \cite{papineni2002bleu}, ROUGE \cite{lin2004rouge}, METEOR \cite{banerjee2005meteor}, and chrF \cite{popovic2015chrf}. Embedding-based metrics such as BERTScore \cite{zhang2020bertscore}, MoverScore \cite{zhao2019moverscore}, and BLEURT \cite{sellam2020bleurt} capture semantic similarity but fail to distinguish logical entailment from contradiction.

Recent work explores LLM-as-a-judge evaluation \cite{zheng2023judgelm,gu2026survey,chiang2023canllmsjudge}, but these methods introduce opacity and instability. Judgements from LLMs are also version dependent. Behind-the-scenes API updates by model providers can completely alter the behavior of the judge over time,  destroying the reproducibility \cite{gu2026survey}. Another study suggests that treating LLM-as-a-judge on 11 real-world text classification and rating tasks, the models arbitrarily resolve their internal indeterminacy, selecting suboptimal, highly unstable outcomes that fail to match human consensus \cite{guerdan2026validating}.

To address these limitations, the proposed \textbf{FOLBench} evaluation suite considers logical coherence, fluency, correctness, and  structural consistency as non-trivial. Briefly, First-Order Logic (FOL) is a formal mathematical system used to represent statements and their relationships \cite{smullyan1968first}. Unlike propositional logic, which can only look at whole sentences as true or false, FOL breaks sentences down into objects, properties, and relations. 
FOL translates natural language into precise symbols:
\begin{itemize}[noitemsep]
    \item Objects (Constants): Specific entities like "John" (\(j\)) or "the book" (\(b\)).
    \item Properties and Relations (Predicates): Characteristics or actions such as \(Person(j)\) or \(Reads(j, b)\).
    \item Quantifiers: Words of scale, specifically \(\forall \) ("for all") and \(\exists \) ("there exists").
    \item Logical Connectives: Links like \(\land \) (AND), \(\lor \) (OR), \(\neg \) (NOT), \(\rightarrow \) (IF... THEN) and \(\leftrightarrow\) (IFF).
\end{itemize}
FOLBench evaluates a candidate FOL statement against a reference FOL statement using a symmetric set of graph-oriented metrics. 

\section{Background and Related Work}
In numerous use cases dependent on LLMs, the output of LLMs is governed and restricted to produce expected output formats \cite{liu2024we}. Since the outputs are in expected formats and at times limited to making choices, evaluation can be done using classification metrics such as Precision, Recall and Fn scores or custom metrics involving rating scales or score ranges\cite{hu2024unveiling, harris2024evaluating}. For other use cases where true text generation is required, existing metrics can fall short of measuring the logical correctness and coherence of the generated text.

\subsection{Surface-Form Metrics}
BLEU \cite{papineni2002bleu}, ROUGE \cite{lin2004rouge}, METEOR \cite{banerjee2005meteor}, and chrF \cite{popovic2015chrf} evaluate text based on lexical surface similarity—meaning they match exact character or word sequences rather than understanding the underlying context or logic. Because they rely on strict string-matching against a human reference text, they struggle to capture structural changes that can alter meaning, logical contradictions, or equivalent phrasing.

\subsection{Embedding-Based Metrics}
Embedding-based metrics such as BERTScore \cite{zhang2020bertscore}, MoverScore \cite{zhao2019moverscore}, and BLEURT \cite{sellam2020bleurt} represent a major leap forward from BLEU and ROUGE by recognizing synonyms and flexible phrasing they still fundamentally fail to judge logical interpretation \cite{anschutz2023not}. These metrics treat meaning as a continuous vector, making them insensitive to discrete logical distinctions.

Surface-Form Metrics and Embedding-based Metrics fail to identify the logical difference and interpretation in the following pair of statement cases. Since metrics try to maximize on token overlaps and ignore the change in meaning due to reordering or missing or addition of token, the logical interpretation is diluted \cite{oliva2025illusion,anschutz2023not}. Such instances of overlooking logical errors can lead to erroneous interpretations \ref{sec:logicalerrors}.

\subsection{Graph-Based Semantic Metrics}
Graph-based evaluation metrics such as Smatch \cite{cai2013smatch}, map sentences to Abstract Meaning Representations (AMR). AMR represents text as a directed acyclic graph where nodes are concepts and edges are semantic roles. AMR handles negation by attaching a \emph{:polarity} - attribute to a specific node in the graph. However, graph alignment metrics such as Smatch evaluate structural similarity by calculating the maximum overlap between nodes and edges. If two graphs are 95\% structurally identical, but one graph features a single missing \emph{:polarity} - node, the Smatch score will remain exceptionally high (e.g., 0.95).

\subsection{LLM-Based Evaluation Frameworks}
LLM-based evaluators \cite{zheng2023judgelm,chiang2023canllmsjudge,xu2025deconstructing} correlate with human preferences but suffer from prompt sensitivity and self-agreement bias. LLM judges also suffer from positional bias between choices and give preferences to verbose outputs over logical correctness \cite{zheng2023judging}. LLMs struggle to distinguish between subtle changes in logical quantifiers. For example, it might judge the sentence 'Almost all patients recovered' as logically equivalent to 'Every patient recovered,' treating a probabilistic statement as an absolute fact \cite{karim2025lost}.

Recent methods have evolved away from simple LLM-as-a-Judge and have pushed the boundaries by introducing strategic frameworks to tackle the problem of open-ended responses. Traditional Multiple Choice Questions (MCQs) suffer from selection bias and random guessing/lottery ticket choices, which mask a model's true limitations. The Open-Style Question-Bench (OSQ-Bench) \cite{myrzakhan2024open} framework addresses this by converting MCQs into open-style questions. Since OSQ-Bench relies on LLM-as-a-Judge, as established in the above sections, it suffers from surface fluency and positional and lengthy bias. MATEval \cite{li2024mateval}introduces multiple LLM personas that debate, reflect, and build a consensus report. While it minimizes individual agent noise, it fails logical validation due to the echo chamber group-thinking, where the agents can overlook and cascade the incorrect reasoning. By prioritizing utility and plausibility within real-world application domains, the framework \cite{ashqar2025benchmarking} allows "good-sounding" text to override logical validity. LogicLLama \cite{yang2024harnessing} is a fine-tuned LLM that translates and corrects Natural Language to First-Order Language conversion. The study introduces the Logical Equivalence (LE) metrics to evaluate the First-Order Language formula by checking whether the ground-truth logic formula (\(\phi \)) and the predicted logic formula (\(\psi \)) logically imply each other. LE recognizes that \(A \land B\) is mathematically identical to \(B \land A\). Primarily, this captures the commutative property but suffers when variable references are not the same or predicates are different.

To that end, FOLBench addresses the opportunity to evaluate open-ended responses on multiple logical avenues. It builds on existing work \cite{yang2024harnessing} and introduces additional metrics to provide a comprehensive logical equivalence.
\section{The FOLBench Framework}
\begin{figure*}
\centering
  \includegraphics[width=0.9\linewidth]{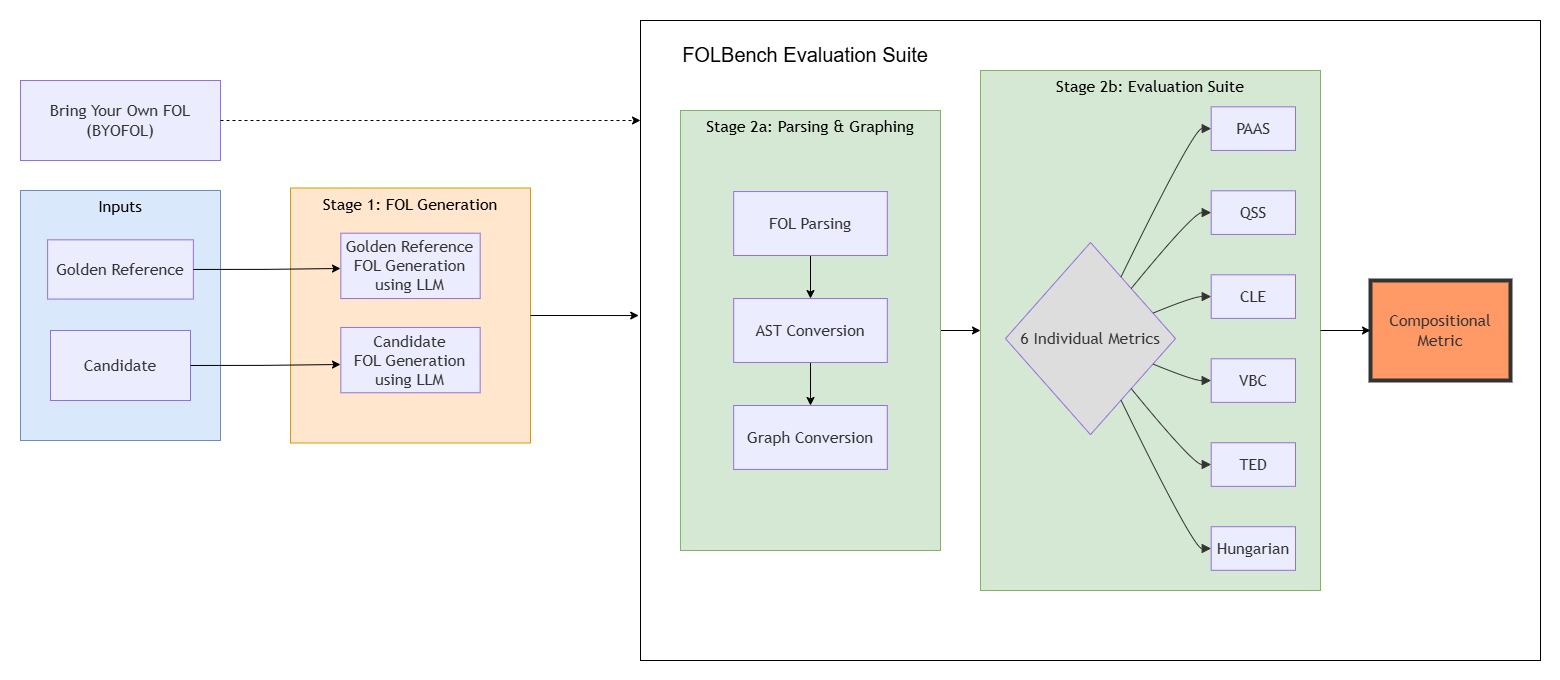}
  \caption{FOLBench Evaluation Framework consists of two stages with Stage 1 focused on preprocessing the inputs that feeds into Stage 2 - Parsing FOLs and graph construction thereby calculating metrics.}
  \label{fig:framework}
\end{figure*}

In contrast to existing metrics and frameworks, FOLBench Evaluation expects a set of preprocessed inputs - First-Order Logic (FOL) formula. The users can either directly feed their own FOL or use LLMs/SLMs to generate FOLs. To reduce the friction for adopting the framework, SLMs \cite{gosuddin_kamaruddin_siddiqi_2026_merged,gosuddin_kamaruddin_siddiqi_2026} are fine-tuned to help generate FOLs equivalents of the corresponding Natural Language candidates.
Broadly, FOLBench (Figure \ref{fig:framework}) consists of two stages: 
\begin{itemize}[noitemsep]
    \item Stage 1: for a given pair of NL convert to FOL or Bring Your Own FOL (BYOFOL),
    \item Stage 2:
    \begin{itemize}
        \item a): Each of the FOLs is parsed, an Abstract Syntax Tree (AST) is generated, and a corresponding network graph is constructed,
        \item b): Using these network graphs, a set of 6 metrics are calculated alongside a weighted compositional metric.
    \end{itemize}
\end{itemize}
\subsection{Stage 1: NL→FOL Conversion Model}
LogicLlama \cite{yang2024harnessing} presents the foundational NL→FOL fine-tuned models with Llama 7B and Llama 13B as base models. These are fine-tuned in two setups, 1) direct translational and 2) naive correction with dependence on initial GPT 4 outputs. LogicLlama models are fine-tuned on the MALLS dataset that is derived from the FOLIO dataset. These relatively large fine-tuned models presents opportunities for training comparatively smaller language models from Gemma 3n 4B and Phi 3.5 4B family. These SLMs are fine-tuned using LoRA on the same FOLIO dataset and quantized using Activation-Aware Weight Quantization (AWQ) \cite{lin2024awq} quantization method to further reduce the required memory footprint. The performance of the models is presented in the Table \ref{tab:model_performance}. The purpose of these models is to reduce friction and increase the adoption of the framework by lowering inference cost and memory requirements. Using LogicLlama performance as the baseline, the quantized version of Phi 3.5 mini 4B performs the best with trade-off for reduced memory footprint.

\begin{table}[htbp]
  \centering
  
    \begin{tabular}{lcc}
    \toprule \toprule
    Model & \multicolumn{2}{c}{\textbf{FOLIO}} \\
    \cmidrule(lr){2-3} 
     & BLEU & LE  \\
    \midrule
    GPT-4 Judge & 0.400 & 0.855 \\
    LogicLlama 7B & 0.385 & 0.849\\
    \textbf{phi 3.5 4B AWQ} & 0.282 & \textbf{0.788} \\
    phi 3.5 4B & 0.169 & 0.635  \\
    \bottomrule
    
    \end{tabular}
    \caption{highlights comparable performance of fine-tuned phi 3.5 4B AWQ quantized model against GPT-4 and LogicLlama 7B}
    \label{tab:model_performance}
\end{table}

\subsection{Stage 2 a: FOL Parsing and Graph Construction}
FOLBench evaluates the logical correctness of First‑Order Logic (FOL) expressions by converting them into a structured representation that supports fine‑grained comparison. The pipeline consists of three stages: (1) parsing the raw FOL string, (2) constructing an abstract syntax tree (AST), and (3) generating a canonical graph representation. All FOLBench metrics operate on this graph, enabling robust evaluation of logical structure beyond surface‑level string similarity.

\subsubsection{FOL Parser}
The FOL parser is a deterministic, grammar‑driven component that transforms a raw FOL string into a typed symbolic structure. It is implemented using a recursive‑descent strategy and enforces the grammar used throughout FOLBench.
The parser first performs lexical analysis by tokenizing the input into:
\begin{itemize}[noitemsep]
    \item  quantifiers (\(\forall \), \(\exists \))
    \item  logical operators (\(\land\), \(\lor\), \(\neg\), \(\longrightarrow\), \(\leftrightarrow\))
    \item  predicate identifiers
    \item  variables and constants
    \item  parentheses and commas
\end{itemize}
Tokens are normalized (e.g., “forall” → \(\forall \)) to ensure consistent downstream processing.
The parser then performs syntactic analysis by validating the formula's structure and constructs a hierarchical representation. It ensures correct operator precedence, properly nested parentheses, well‑formed predicate argument lists, valid quantifier placement, and no illegal operator sequences. Malformed formulas are rejected. To ensure consistent evaluation, the parser normalizes variable names (e.g., x1, x2, …), predicate argument ordering, connective precedence, and redundant parentheses. This normalization ensures that logically equivalent formulas with different surface forms produce identical ASTs.

\subsubsection{Abstract Syntax Tree}
The AST is a hierarchical representation of the logical formula produced by the parser. Each node corresponds to a semantic unit with quantifier nodes (e.g., \(\forall \)x, \(\exists \)y), predicate nodes (e.g., Student(x), Loves(x,y)), logical operator as nodes, variable nodes, and constant nodes.
The AST structure is essential for detecting logical errors such as quantifier inversion, variable capture, argument swaps, missing conjuncts and incorrect negation scope. The AST is the direct input to the graph builder.

\subsubsection{Graph Construction}

The AST is transformed into a canonical directed graph that captures the full logical structure of the formula. This graph is the foundation for all FOLBench metrics. Graph Nodes correspond to predicates, variables, constants, quantifiers and logical operators. Graph Edges encode semantic relationships:
\begin{itemize}[noitemsep]
    \item  predicate‑argument edges connects predicate nodes to their argument nodes.
    \item  quantifier‑binding edges connects quantifier nodes to the variables they bind.
    \item  operator edges connects logical operators to their operands
    \item  scope edges represent the hierarchical structure of quantifier domains
\end{itemize}
The graph builder applies normalization rules to ensure structural consistency by normalizing commutative operators (\(\land\), \(\lor\)), flattening nested conjunctions/disjunctions and consistent ordering of predicate arguments as in ASTs

\subsection{Stage 2b: Graph-Based Metrics}
FOLBench evaluates logical correctness by comparing the canonical graphs of a candidate formula and its gold reference. Each metric targets a specific aspect of logical structure—equivalence, shape, quantifier behavior, variable binding, predicate–argument alignment, and optimal node matching via Hungarian similarity. Together, they provide a rich, interpretable profile of where and how a candidate logic expression deviates from the reference.
FOLBench evaluates by comparing the graphs of a candidate formula \(G_c\) and a gold formula \(G_g\). All metrics return values in \([0,1]\). Let:
\begin{itemize}[noitemsep]
    \item \(V_c, E_c\) be the nodes and edges of the candidate graph
    \item \(V_g, E_g\) be the nodes and edges of the gold graph
\end{itemize}

\subsubsection{Predicate-Argument Alignment Score (PAAS)}
PAAS is the main metric for fine‑grained relational correctness. PAAS evaluates whether predicates and their arguments are aligned correctly between candidate and gold graphs. It does so by comparing predicate nodes, argument positions such as subject, object, and the mapping between predicate–argument structures in both graphs. It computes a similarity score based on matched predicates with correct arguments, penalizing argument swaps, missing arguments, or extra arguments, and normalizing by the number of predicate–argument pairs in the gold graph. PAAS helps uncover
\begin{itemize}[noitemsep]
    \item argument swaps (Parent\emph{(Alice, Bob)} vs Parent\emph{(Bob, Alice)})
    \item hallucinated arguments (P(x) vs P(x) \(\land\) \emph{R}(x))
    \item missing arguments or predicates
    \item wrong predicate usage with correct entities
\end{itemize}
\(P_g\) be the set of predicate nodes in the gold graph. \({args}(p)\) be the argument list of predicate \(p\)
Define argument alignment as:
$$ArgAlign(p_c, p_g) = \frac {| args(p_c) \cap args(p_g)|}{args (p_g) |}$$
then,
\begin{equation}
    \begin{split}
        {PAAS} (G_c, G_g) = \\
    \frac{1}{|P_g|}
    \sum_{p_g \in P_g} \
    \max_{p_c \in P_c}{ArgAlign}(p_c, p_g)
    \end{split}
\end{equation}
\subsubsection{Quantifier Scope Similarity (QSS)}
Quantifier Scope Similarity (QSS)
QSS is crucial for detecting errors that radically change meaning while leaving most tokens intact. QSS evaluates whether quantifiers in the candidate formula have the same scope and structure as in the gold formula. QSS compares the set of quantifiers (\(\forall \), \(\exists\)), their bound variables, and the scope of the subgraphs. It measures overlap between quantifier–scope structures in the two graphs, by matching quantifier nodes, comparing their descendant subgraphs, and computing a normalized similarity score. QSS excels at catching \(\forall \) vs \(\exists\) swaps, scope inversion (e.g., \(\forall \)x \(\exists\)y vs \(\exists\)y \(\forall \)x), missing or extra quantifiers, and quantifiers attached to the wrong sub-formula.
Let, \(Q_c, Q_g\) be the sets of quantifier nodes and \({scope}(q)\) be the set of nodes under quantifier \(q\).
Define quantifier alignment as:
\begin{equation}
{Align}(q_c, q_g) =
\frac{|{scope}(q_c) \cap {scope}(q_g)|}
{|{scope}(q_g)|}    
\end{equation}
then,
\begin{equation}
    \text{QSS}(G_c, G_g) =
\frac{1}{|Q_g|}
\sum_{q_g \in Q_g}
\max_{q_c \in Q_c} \text{Align}(q_c, q_g)
\end{equation}
\subsubsection{Variable Binding Consistency (VBC)}
VBC is vital for proofs and complex quantified formulas. VBC checks whether variables are bound and used correctly relative to the gold formula. VBC inspects which quantifier binds each variable, where each variable appears, and whether any variable is captured incorrectly or left free. It compares binding structures between candidate and gold graphs and returns a normalized score based on correctly bound variables, correctly scoped occurrences, and absence of illegal captures or free variables. VBC identifies variable capture (\(\forall x\) \(\exists x\) vs \(\forall x\) \(\exists y\)), free variables that should be bound, variables bound by the wrong quantifier and missing or extra variable occurrences.
Let \(bind(v)\) be the quantifier that binds variable \(v\),
\begin{equation}
VBC(G_c, G_g) =
\frac{|\{v \in V_g : bind_c(v) = bind_g(v)\}|}
{|V_g|}  
\end{equation}
Intuitively, VBC checks the association between quantifiers and variables, while QSS checks the territory or span each quantifier governs.
\subsubsection{Canonicalized Logical Equivalence (CLE)}
CLE is the strongest signal that a candidate formula is “semantically correct” despite surface differences. CLE detects whether two formulas are logically equivalent, even if their surface forms differ. CLE is a binary or graded score indicating whether the candidate and gold graphs represent the same logical function under canonicalization. The metric applies normalization rules (e.g., De Morgan, commutativity of \(\land\)/\(\lor\), double negation), compares canonical graphs for structural identity and returns 1.0 for equivalent formulas, 0.0 for non‑equivalent, or an intermediate score if partial equivalence is used. It identifies equivalent rewrites (e.g., \(\neg\)(P \(\land\) Q) vs \(\neg\)P \(\lor\) \(\neg\)Q), variable renaming, and reordered but logically identical conjuncts/disjuncts. 
Let \(\mathcal{C}(\cdot)\) be the canonicalization function.
\begin{equation}
CLE(G_c, G_g) =
\begin{cases}
1, & \text{if } \mathcal{C}(G_c) = \mathcal{C}(G_g) \\
0, & \text{otherwise}
\end{cases}
\end{equation}
\subsubsection{Tree Edit Distance (TED)}
TED measures how much structural change is required to transform the candidate graph into the gold graph. TED computes the minimum cost sequence of node insertions, deletions, and substitutions needed to transform the candidate’s tree (or tree‑like graph) into the reference \cite{zhang1989simple}. TED catches missing predicates or operators, extra (hallucinated) predicates, wrong connective types (e.g., \(\land\) vs \(\lor\) vs \(\rightarrow\)), and large structural deviations in formula shape. TED provides a global view of structural similarity, complementing CLE’s equivalence check.
Let \(\text{EditCost}(G_c, G_g)\) be the minimum edit cost, \(N = \max(|V_c|, |V_g|)\)
\begin{equation}
\text{TED}(G_c, G_g) = 1 - \frac{\text{EditCost}(G_c, G_g)}{N}    
\end{equation}
\subsubsection{Hungarian similarity (HS)}
Hungarian Similarity provides a holistic view of how well the candidate graph can be aligned to the gold graph under optimal matching. HS finds the optimal node‑to‑node matching between candidate and gold graphs to quantify overall structural and semantic alignment. Hungarian Similarity uses the Hungarian algorithm \cite{kuhn1955hungarian} to solve a bipartite matching problem between nodes in the candidate and gold graphs:
\begin{itemize}[noitemsep]
    \item Construct a cost matrix where each entry represents the dissimilarity between a candidate node and a gold node (based on type, label, role, etc.).
    \item Run the Hungarian algorithm to find the minimum‑cost matching.
    \item Convert the total matching cost into a normalized similarity score.
\end{itemize}
HS identifies best possible alignment even when graphs differ in size as it focuses on global structural similarity beyond local matches,  subtle mismatches in node roles (e.g., predicate vs operator) and cases where local metrics (PAAS, QSS) are partially correct but global alignment is poor.
Let \(M\) be the cost matrix where \(M_{ij}\) is the dissimilarity between candidate node \(v_{c,i}\) and gold node \(v_{g,j}\). \({Hungarian}(M)\) returns the minimum total matching cost and \(C_{\max}\) is the maximum possible cost
\begin{equation}
    \text{HS}(G_c, G_g) =
1 - \frac{{Hungarian}(M)}{C_{\max}}
\end{equation}
Intuitively, HS is like aligning two sentences by matching each word to the most similar word in the other sentence. TED is like comparing two sentences by counting how many words you must insert/delete/replace.

\section{Experiments}
AMR-LE \cite{bao-etal-2024-abstract} and ConjNLI \cite{saha2020conjnlinaturallanguageinference} datasets are examples of similar but logical variant NL pairs. The sensitivity of FOLBench metrics is evaluated across the train set of AMR-LE (Logical Non/Equivalence) and the test set of ConjNLI (Entailment, Neutral, and Contradiction). Sentence pairs from both benchmarks are first translated into FOL representation via the fine-tuned LLM, and the resulting derivations are scored using FOLBench. To examine how metric profiles correspond to logical classes, the metric vectors are clustered per label using DBSCAN.
The resulting clusters reveal distinct groupings that mirror the ground-truth NLI and equivalence labels. Detailed qualitative analyses of representative cluster exemplars—illustrating how individual FOLBench metrics capture specific logical transformations—are provided in Appendix~\ref{sec:appendix b} .

\section*{Conclusion}
FOLBench, is a logic-grounded benchmark for evaluating open-ended LLM responses via fine-tuned NL→FOL conversion and graph-based semantic metrics. FOLBench offers interpretable, robust, and semantically precise evaluation, revealing reasoning failures invisible to existing metrics. As the content generated by LLMs continue to explode, such evaluation will be essential for ensuring reliability and semantic fidelity.

\section*{Limitations}
FOLBench opens new avenues for grounding logical correctness, such evaluation metrics are in nascent stages and should continue to evolve and mature. Even though the metrics are reproducible and deterministic, the supply FOL can be improved as the generation of FOL remains susceptible LLMs hallucinations. On the other hand, human-generated FOLs are expensive to produce. Current SLMs are generic and can benefit from domain-based fine-tuning for better adoption.

\bibliography{latex/custom}

\section*{Appendix}
\subsection{Appendix A: Types of Logical Errors}
\label{sec:appendix a}
\begin{figure*}[t]
\centering
  \includegraphics[width=0.9\linewidth]{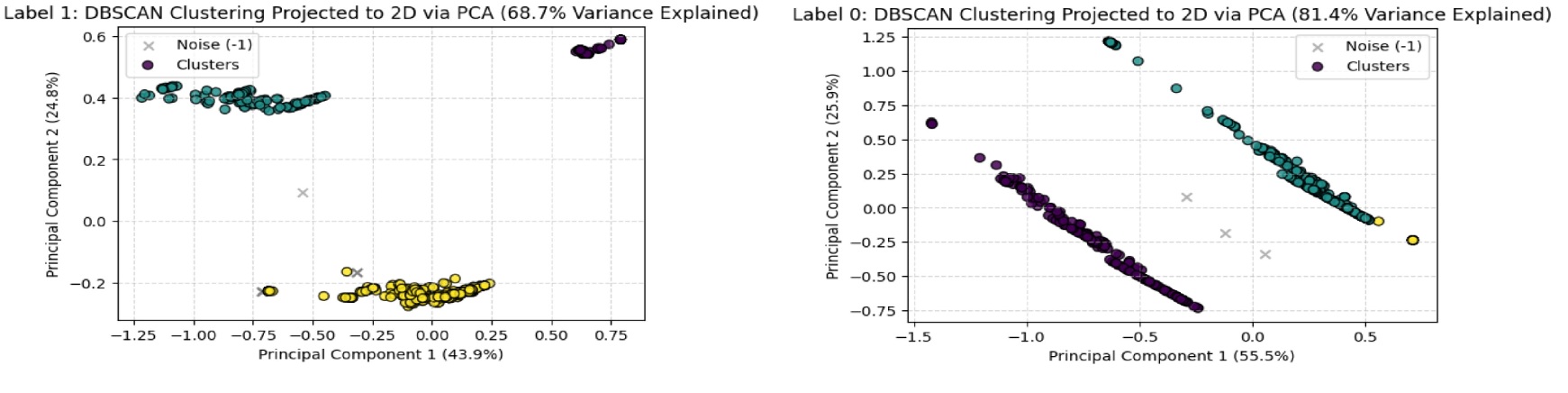}
  \caption{Cluster Representation of metrics for Label 1 and 0.}
  \label{fig:amr_le_sbs}
\end{figure*}
\begin{figure*}[t]
\centering
  \includegraphics[width=0.9\linewidth]{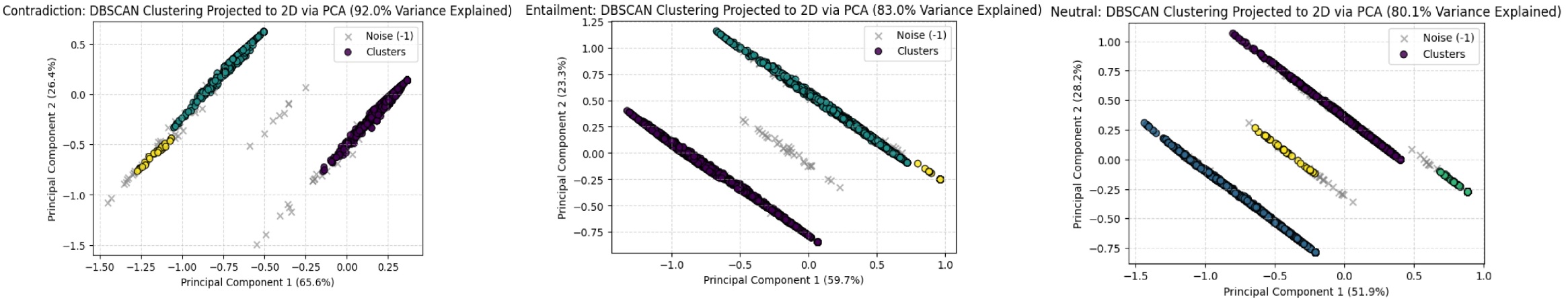}
  \caption{ConjNLI Cluster Representation of metrics for Label Contradiction, Entailment and Neutral.}
  \label{fig:conj_nli_sbs}
\end{figure*}
\label{sec:logicalerrors}
Set of logical errors, overlooked and not accounted for by existing metrics:
\begin{itemize}[noitemsep]
    \item Negation flip
    \begin{itemize}[noitemsep]
        \item Sentence 1: "Every bird \emph{can} fly"
        \item Sentence 2: "Every bird \emph{cannot} fly"
    \end{itemize}
    \item Quantifier swap
    \begin{itemize}[noitemsep]
        \item Sentence 1: "\emph{Everyone} loves \emph{someone}"
        \item Sentence 2: "\emph{Someone} loves \emph{everyone}"
    \end{itemize}
    \item Implication vs Conjunction
    \begin{itemize}[noitemsep]
        \item Sentence 1: "If it rains, the ground gets wet"
        \item Sentence 2: "It rains \emph{and} the ground gets wet"
    \end{itemize}
    \item Argument swap
    \begin{itemize}[noitemsep]
        \item Sentence 1: "\emph{Alice} is \emph{Bob’s} parent"
        \item Sentence 2: "\emph{Bob} is \emph{Alice’s} parent"
    \end{itemize}
    \item Variable binding error
    \begin{itemize}[noitemsep]
        \item Sentence 1: "Every student admires \emph{some} teacher"
        \item Sentence 2: "Every student admires \emph{every} teacher"
    \end{itemize}
    \item Wrong logical operator
    \begin{itemize}[noitemsep]
        \item Sentence 1: "The light is on \emph{or} the fan is running"
        \item Sentence 2: "The light is on \emph{and} the fan is running"
    \end{itemize}
\end{itemize}

\subsection{Appendix B: Experiment Results}
\label{sec:appendix b}
\subsubsection{AMR-LE Dataset}

The dataset\footnote{Dataset: https://github.com/Strong-AI-Lab/Logical-Equivalence-driven-AMR-Data-Augmentation-for-Representation-Learning/} is available on the repository associated with the paper \cite{bao-etal-2024-abstract}. Each pair of sentence are labeled either as Logically Equivalent - Label 1 or Logically Nonequivalent - Label 0. Of the 2,836 rows, 33 rows are manually excluded after the FOL representation due to hallucination. FOLBench metrics are ran for the remainder of the dataset. For both the labels, clustering is performed using DBScan on the vectors constructed using FOLBench metrics (excluding the composite score). To better visualize, first 2 Principal Components are extracted and presented in the following figure \ref{fig:amr_le_sbs}

Representative examples for Logically Equivalent Data:
\begin{itemize}[noitemsep]
    \item Cluster 1
    \begin{itemize}[noitemsep]
        \item Sentence 1: "If Anne is not big, then the dinosaur is not powerful."
        \item Sentence 2: "If the dinosaurs were powerful, Anne would be big."
    \end{itemize}
    \item Cluster 2
    \begin{itemize}[noitemsep]
        \item Sentence 1: "If Alan is little, then the crocodile is not small."
        \item Sentence 2: "Alan isn't little or a crocodile isn't small."
    \end{itemize}
    \item Cluster 3
    \begin{itemize}[noitemsep]
        \item Sentence 1: "the rabbit is not beautiful and Dave is not round."
        \item Sentence 2: "Dave isn't round, and the bunny isn't beautiful."
    \end{itemize}
\end{itemize}

Representative examples for Logically Non-Equivalent Data:
\begin{itemize}[noitemsep]
    \item Cluster 1
    \begin{itemize}[noitemsep]
        \item Sentence 1: "If Fiona is not bad, then Charlie is quiet."
        \item Sentence 2: "Fiona wasn't bad or Charlie quiet."
    \end{itemize}
    \item Cluster 2
    \begin{itemize}[noitemsep]
        \item Sentence 1: "If the dinosaur is horrible, then the leopard is tiny."
        \item Sentence 2: "The dinosaurs weren't horrible or the leopard wasn't tiny."
    \end{itemize}
    \item Cluster 3
    \begin{itemize}[noitemsep]
        \item Sentence 1: "If the lion is not strong, then the crocodile is not quiet."
        \item Sentence 2: "The crocodiles would not be quiet if the lions were strong."
    \end{itemize}
\end{itemize}

\subsubsection{ConjNLI Dataset}

The dataset\footnote{Dataset: https://github.com/swarnaHub/ConjNLI/tree/master/data/NLI} is available on the repository associated with the paper \cite{saha2020conjnlinaturallanguageinference}. Each pair of sentence are labeled Contradiction, Entailment, and Neutral. For ConjNLI, training dataset is used for evaluation to add volume of sentence pairs to the experiment. Of the 15,000 rows, 488 rows are manually excluded after the FOL representation due to hallucination. FOLBench metrics are ran for the remainder of the dataset. For both the labels, clustering is performed using DBScan on the vectors constructed using FOLBench metrics (excluding the composite score). To better visualize, first 2 Principal Components are extracted and presented in the following figure \ref{fig:conj_nli_sbs}

Representative examples from Contradiction Data:
\begin{itemize}[noitemsep]
    \item Cluster 1
    \begin{itemize}[noitemsep]
        \item Sentence 1: "Proud Mary is an upcoming American action-thriller directed by Babak Najafi, from a screenplay written by John S. Newman and Christian Swegal."
        \item Sentence 2: "Proud Mary is an upcoming American action-thriller directed by Babak Najafi, from a screenplay written by John S. Newman and Euphemia Duncan."
    \end{itemize}
    \item Cluster 2
    \begin{itemize}[noitemsep]
        \item Sentence 1: "The Simpsons focuses on the eponymous family, which consists of Homer, Marge, Bart, Lisa and Maggie."
        \item Sentence 2: "The Simpsons focuses on the eponymous family, which consists of Homer, Marge, Bart, Lisa and Grossman."
    \end{itemize}
    \item Cluster 3
    \begin{itemize}[noitemsep]
        \item Sentence 1: "What occurred then is unclear but despite a meeting between Verginius and Vindex, the forces under Verginius seemed to have decided on a battle without orders."
        \item Sentence 2: "despite a meeting between Verginius and Vindex, the forces under Verginius seemed to have decided on a battle without orders."
    \end{itemize}
\end{itemize}

Representative examples from Entailment Data:
\begin{itemize}[noitemsep]
    \item Cluster 1
    \begin{itemize}[noitemsep]
        \item Sentence 1: "An acoustic version of the song does appear on YouTube but is not featured on the Story Untold EP."
        \item Sentence 2: "An acoustic version of the song does appear on YouTube."
    \end{itemize}
    \item Cluster 2
    \begin{itemize}[noitemsep]
        \item Sentence 1: "A test that could but does not run contrary to the hypothesis corroborates the theory."
        \item Sentence 2: "A test that could corroborates the theory."
    \end{itemize}
    \item Cluster 3
    \begin{itemize}[noitemsep]
        \item Sentence 1: "He was billed as the brother of Fritz Von Erich, but was not related to Fritz or the rest of the Von Erich family."
        \item Sentence 2: "He was billed as the brother of Fritz Von Erich, but was not related to the rest of the Von Erich family."
    \end{itemize}
\end{itemize}

Representative examples from Neutral Data:
\begin{itemize}[noitemsep]
    \item Cluster 1
    \begin{itemize}[noitemsep]
        \item Sentence 1: "The Ezhou - Huanggang built - up " " area was home to 1,035,496 inhabitants from the Echeng and Huangzhou, Huanggang Districts."
        \item Sentence 2: "The Ezhou - Huanggang built-up ("or metro") area was home to 1,035,496 inhabitants from the Echeng and Huangzhou, Huanggang Districts."
    \end{itemize}
    \item Cluster 2
    \begin{itemize}[noitemsep]
        \item Sentence 1: "It was the first x86 processor designed by AMD from the ground up, and not reverse - engineered as previous generations of x86 processors produced by AMD."
        \item Sentence 2: "It was the first x86 processor designed by AMD from the ground up, and not licensed or reverse-engineered as previous generations of x86 processors produced by AMD."
    \end{itemize}
    \item Cluster 3
    \begin{itemize}[noitemsep]
        \item Sentence 1: "The majority of short stories which comprise the book, were written between 1993 - 2004 with two stories having been known to exist as early as 1945."
        \item Sentence 2: "The majority of short stories which comprise the book, were written between 1993-2004 with one or two stories having been known to exist as early as 1945."
    \end{itemize}
\end{itemize}

\end{document}